# On revising fuzzy belief bases

Richard Booth and Eva Richter
University of Leipzig
Department of Computer Science
Augustusplatz 10/11
04109 Leipzig, Germany
{booth,erichter}@informatik.uni-leipzig.de

## Abstract

We look at the problem of revising fuzzy belief bases, i.e., belief base revision in which both formulas in the base as well as revision-input formulas can come attached with varying truth-degrees. Working within a very general framework for fuzzy logic which is able to capture certain types of uncertainty calculi as well as truth-functional fuzzy logics, we show how the idea of rational change from "crisp" base revision, as embodied by the idea of partial meet (base) revision, can be faithfully extended to revising fuzzy belief bases. We present and axiomatise an operation of partial meet *fuzzy base* revision and illustrate how the operation works in several important special instances of the framework.

## 1 INTRODUCTION

The ability to rationally change one's beliefs in the face of new information which, possibly, contradicts the currently held beliefs is a basic characteristic of intelligent behaviour. Hence the question of belief revision is an important question in AI. A very successful framework in which this question is studied is the one due to Alchourrón, Gärdenfors and Makinson (AGM) [1, 4], with its operation of *partial meet revision*. One limitation of this framework is that belief in a formula is taken as a matter of all or nothing: either the formula is believed or it is not. However, real-life knowledge bases may well contain information of a more *graded* nature. For instance we might want to represent information about *vague* concepts or *uncertain* beliefs. Likewise revision inputs may come with a degree attached. Our aim in this paper is to examine revision in the general setting which allows for such different degrees, while keeping the spirit of AGM.

As a most suitable backdrop in which to work out our ideas we choose a very general framework for *fuzzy logic* due to Gerla [7]. The basic construct here is that of an *abstract fuzzy deduction system*, which generalises Tarski's notion of deductive systems. Roughly, this consists of three basic ingredients: *(i)* a set $L$ of formulas to describe the world, *(ii)* a set $W$ of *truth-degrees* (whose precise interpretation is mostly left open) which may be assigned to the formulas to create *fuzzy belief bases*, and *(iii)* a *fuzzy deduction operator* $D$ which takes as input a fuzzy base $u$ and returns another fuzzy base $D(u)$ representing its *(fuzzy) conclusions*. Sometimes a fourth ingredient is included – a *fuzzy semantics* $\mathcal{M}$ – in which case we speak of an *abstract fuzzy logic*. When $W = \{0, 1\}$ we find ourselves in the usual "crisp" setting of AGM. The framework has also been shown capable of capturing several different flavours of uncertain reasoning, including truth-functional logic and certain types of probabilistic logic.

Within this fuzzy framework, the question of revision we are interested in then takes the following form: Given a fuzzy base $u$ representing our current information, how should we change $u$ to incorporate the new information that the truth-degree of some formula $\phi$ is at least $a$ for some $a \in W$? In this paper we assume that the object of change $u$ is an arbitrary fuzzy base which need not be deductively closed, i.e., possibly $u \neq D(u)$. Indeed, following the often referred-to distinction made in the belief revision literature (e.g. [13, p. 22]), our approach will be *foundationalist* rather than *coherentist*. That is, we differentiate between those beliefs which are "basic" or "explicit", ($u$) and those which are "merely derived" or "implicit" (i.e., that information in $D(u)$ which goes strictly beyond that contained in $u$).

The original AGM theory was a theory about how to revise deductively closed sets of formulas, but the more general case of revising arbitrary (crisp) bases has also been studied, notably by Hansson [11, 13], who axiomatically characterised partial meet *base* re-vision. We will generalise this operation into *partial meet fuzzy base revision* and give an axiomatisation. Surprisingly, despite the increase in complexity which admitting many truth-degrees brings, the form of the axiomatisation is roughly the same as in the crisp case. This shows how the principles on which partial meet revision are based really require very little structure. The set of truth-degrees is not even required to be linearly-ordered – any complete, distributive lattice will do.

The plan of the paper is as follows. In Sect. 2 we set up the framework of abstract fuzzy logic and describe some instances of it, including those related to truth-functional fuzzy logics, necessity logic and probability logic. In Sect. 3 we define partial meet fuzzy base revision operators and give examples to illustrate how these operators work for each instance of the framework from the previous section. We give the axiomatisation of partial meet fuzzy base revision in Sect. 4. Finally we conclude in Sect. 5.

## 2   ABSTRACT FUZZY LOGIC

Our first task is to formally define abstract fuzzy deduction systems. The following definitions are based on [7]. As we said above, we assume $L$ to be the set of all formulas. We take the set $W$ of all possible truth-degrees to be a complete lattice, i.e., we assume $W$ to come equipped with a partial order $\leq_W$ on $W$ such that every $A \subseteq W$ has both a *supremum* (or *join*) $\sup(A)$ and an *infimum* (or *meet*) $\inf(A)$. For $a, b \in W$ we write $a \curlyvee b$ for $\sup(\{a,b\})$ and $a \curlywedge b$ for $\inf(\{a,b\})$. Often (for instance in our examples) $W$ will be linearly ordered (e.g., the real unit interval). However, in general the only additional assumption we make about $W$ is that it is also *distributive*, i.e., that for all $a, b, c \in W$ we have $a \curlywedge (b \curlyvee c) = (a \curlywedge b) \curlyvee (a \curlywedge c)$, equivalently, $a \curlyvee (b \curlywedge c) = (a \curlyvee b) \curlywedge (a \curlyvee c)$.[1] We use $0_W$ and $1_W$ to denote the minimal and maximal elements of $W$.

A *fuzzy belief base* is then just an assignment $u : L \to W$ of truth-degrees to the formulas. Such a piece of information $u$ should be understood as an under constraint, i.e. $u(\phi) = a$ means that the truth-degree of $\phi$ is *at least* $a$. We denote the set of all possible fuzzy bases by $\mathcal{F}(L)$. The ordering $\leq_W$ induces a "fuzzy subset" relation $\sqsubseteq$ on $\mathcal{F}(L)$ by taking, for $u, v \in \mathcal{F}(L)$, $u \sqsubseteq v$ iff $u(\phi) \leq_W v(\phi)$ for all $\phi \in L$. The meaning of this is that $v$ carries more (or more exact) information than $u$. With this definition it is easy to see that $(\mathcal{F}(L), \sqsubseteq)$ forms a complete, distributive lattice. Given $X \subseteq \mathcal{F}(L)$ we shall denote the supremum and infimum of $X$ under $\sqsubseteq$ by $\bigsqcup X$ and $\bigsqcap X$ respectively. We write $u \sqcup v$ for $\bigsqcup\{u,v\}$ and $u \sqcap v$ for $\bigsqcap\{u,v\}$. We have the following, for all $X \subseteq \mathcal{F}(L)$ and $\phi \in L$,

$$\left[\bigsqcup X\right](\phi) = \sup(\{u(\phi) \mid u \in X\})$$

$$\left[\bigsqcap X\right](\phi) = \inf(\{u(\phi) \mid u \in X\})$$

We use $\sqsubset$ to denote the strict part of $\sqsubseteq$. The $\sqsubseteq$-maximal element of $\mathcal{F}(L)$, i.e., the fuzzy base which assigns degree $1_W$ to every formula, will be denoted by $u_\perp$. The $\sqsubseteq$-minimal element of $\mathcal{F}(L)$, i.e., the fuzzy base which assigns degree $0_W$ to every formula, will be denoted by $u_\top$. For a fuzzy base $u$ we call the set of formulas $\phi$ for which $u(\phi) \neq 0_W$ the *support of* $u$ and denote this set by $Supp(u)$. If $Supp(u) = \{\phi_1, \ldots, \phi_k\}$ is finite then we may represent $u$ as $\{(\phi_1/a_1), \ldots, (\phi_k/a_k)\}$ with the interpretation that $u(\phi_i) = a_i$ for $i = 1, \ldots, k$. We will often use $(\phi/a)$ to denote the base $\{(\phi/a)\}$. Although the support of a fuzzy base will typically be finite, the results we describe will be valid for arbitrary $u$.

The tool for drawing conclusions is the fuzzy deduction operator $D : \mathcal{F}(L) \to \mathcal{F}(L)$. It is assumed to satisfy analogues of the three basic Tarski properties:

- $u \sqsubseteq v$ implies $D(u) \sqsubseteq D(v)$    (Monotony)
- $D(D(u)) = D(u)$    (Idempotence)
- $u \sqsubseteq D(u)$    (Reflexivity)

If $D(u) = u_\perp$ then we say that $u$ is *D-inconsistent*, otherwise *D-consistent*. (We omit the "D-" if it is clear from the context.) A *(fuzzy) theory* is any fixed point of $D$. Another property of $D$, which will be important to us, is *logical compactness*:

**Definition 1 ([7])** *Let $D : \mathcal{F}(L) \to \mathcal{F}(L)$ be a deduction operator. Then $D$ is* logically compact *iff we have $D(\bigsqcup X) \neq u_\perp$ for all $X \subseteq \mathcal{F}(L)$ such that (i) $u \in X$ implies $D(u) \neq u_\perp$, and (ii) for all $u, v \in X$ there exists $w \in X$ such that $u \sqcup v \sqsubseteq w$.*

Using an order-theoretical term, the definition says that $D$ is logically compact iff the supremum of every *directed* family of $D$-consistent fuzzy bases is itself $D$-consistent.

We are now able to give the following formal definition:

**Definition 2** *An* abstract fuzzy deduction system *is a triple $(L, W, D)$ where $L$ is a set of formulas, $W$ is a complete, distributive lattice of truth-degrees and $D$ is a logically compact fuzzy deduction operator which satisfies (Monotony), (Idempotence) and (Reflexivity).*

Sometimes (especially for our examples) it is convenient to describe the deduction operator $D$ of an abstract fuzzy deduction system semantically. An *abstract fuzzy semantics* is a subset $\mathcal{M}$ of $\mathcal{F}(L)$, such that $u_\perp \notin \mathcal{M}$, whose elements are called *models*. Intuitively the models represent complete descriptions of "possible worlds", whereas the fuzzy bases $u$ not in $\mathcal{M}$ represent incomplete knowledge. An element $m \in \mathcal{M}$ is a model of a fuzzy base $u$ if $u \sqsubseteq m$. We denote the set of models of $u$ in $\mathcal{M}$ by $\text{mod}_\mathcal{M}(u)$. An abstract fuzzy semantics $\mathcal{M}$ yields a fuzzy deduction operator $J_\mathcal{M}$ by setting, for each $u \in \mathcal{F}(L)$, $J_\mathcal{M}(u) = \bigsqcap \text{mod}_\mathcal{M}(u)$. It is easy to see that $J_\mathcal{M}$ satisfies (Monotony), (Idempotence) and (Reflexivity), and also that a fuzzy base $u$ is $J_\mathcal{M}$-consistent iff $\text{mod}_\mathcal{M}(u) \neq \emptyset$.

---

[1]For another general approach to modelling uncertainty which likewise relaxes the assumption of linearity see [10].

**Definition 3** *An abstract fuzzy logic is a quadruple $(L, W, D, \mathcal{M})$ where $(L, W, D)$ is an abstract fuzzy deduction system and $\mathcal{M}$ is an abstract fuzzy semantics such that $D = J_\mathcal{M}$ (i.e., the "completeness theorem" holds).*

For any abstract fuzzy deduction system we can always associate a suitable semantics: just take $\mathcal{M}$ to be the set of all $D$-consistent theories.

## 2.1 CONCRETE EXAMPLES

We now give a few example instantiations of the above framework. In each of these we take the set of formulas to be the set of formulas $L_{\text{Prop}}$ from a propositional language closed under the connectives $\neg, \land, \lor$ and $\rightarrow$. We treat $\theta \leftrightarrow \phi$ as an abbreviation for $(\theta \rightarrow \phi) \land (\phi \rightarrow \theta)$. We denote the classical logical consequence operator of propositional logic by $Cn$.

### 2.1.1 Crisp Deduction Systems

The simplest example of a set of truth-degrees is, of course, the case when $W$ consists of just two elements $\{0, 1\}$ standing for "false" and "true" respectively. In this case belief bases $u$ are "crisp", i.e., they correspond to (characteristic functions of) sets of formulas in $L_{\text{Prop}}$, and $\sqsubseteq, \sqcap, \sqcup$ effectively reduce to the usual $\subseteq, \cap, \cup$ (thus in this case we write the more usual "$\phi \in u$" rather than "$u(\phi) = 1$" etc.). In the belief revision literature it is customary to assume that, in addition to (Monotony), (Idempotence) and (Reflexivity), the deduction operator $D$ satisfies the following three rules:

- If $\phi \in Cn(u)$ then $\phi \in D(u)$   (Supraclassicality)
- $\phi \in D(u \cup \{\theta\})$ iff $(\theta \rightarrow \phi) \in D(u)$   (Deduction)
- If $\phi \in D(u)$ then $\phi \in D(u')$ for some finite $u' \subseteq u$ (Compactness)

We shall call an abstract fuzzy deduction system of the form $(L_{\text{Prop}}, \{0, 1\}, D)$ where $D$ satisfies the above three properties a *crisp deduction system*. That $D$ is logically compact follows from the following observation, which is easy to verify:

**Proposition 1** *Let $D : 2^{L_{\text{Prop}}} \rightarrow 2^{L_{\text{Prop}}}$ be a deduction operator which satisfies (Supraclassicality) and (Deduction). Then $D$ satisfies (Compactness) iff $D$ is logically compact in the sense of Definition 1.*

Thus we see that, for crisp deduction systems, the property of logical compactness collapses into the usual notion of compactness. Note that for a semantics here we could take $\mathcal{M}$ to consist of all the maximal consistent theories.

### 2.1.2 Łukasiewicz Fuzzy Logic

In the rest of our examples we take $W = [0, 1]$, i.e., the real unit interval equipped with the usual ordering $\leq$. Each example will differ only in the choice of a semantics, i.e., what counts as a "possible world", leading to different types of deduction operator. The first is related to infinitely many-valued Łukasiewicz logic (see, for example [8, 15]). We take as the semantics the set $\mathcal{M}_{\text{luk}}$ of all *truth-functional valuations* over $L_{\text{Prop}}$ in the many-valued Łukasiewicz logic, i.e., the set of functions $m : L_{\text{Prop}} \rightarrow [0, 1]$ satisfying, for all $\theta, \phi \in L_{\text{Prop}}$,

$$\begin{aligned} m(\neg\theta) &= 1 - m(\theta) \\ m(\theta \land \phi) &= m(\theta) \barwedge m(\phi) \\ m(\theta \lor \phi) &= m(\theta) \veebar m(\phi) \\ m(\theta \rightarrow \phi) &= 1 \barwedge (1 - m(\theta) + m(\phi)) \end{aligned}$$

(Note that here "$\rightarrow$" does not behave as material implication.) So here the "fuzziness" arises from having worlds with graded properties. We then take $D_{\text{luk}} = J_{\mathcal{M}_{\text{luk}}}$. It can be shown [15, Lemma 4.17] that for any given fuzzy base $u$ we have

$u \sqcup (\phi/a)$ is inconsistent iff $D_{\text{luk}}(u)(\neg\phi) > 1 - a$   (*)

We also have the following:

**Proposition 2 ([7])** $D_{\text{luk}}$ *is logically compact.*

For an example of a fuzzy base in this logic let $x, y, z$ be distinct propositional variables and consider:

$$u_0 = \{(x/0.75), (x \rightarrow y/0.75), (z/0.25)\}.$$

For an example of an inference we have $D_{\text{luk}}(u_0)(y) = 0.5$, i.e., we infer that the truth-degree of $y$ is at least 0.5. To see this, we have

$$D_{\text{luk}}(u_0)(y) = \inf\{m(y) \mid m \in \text{mod}_{\mathcal{M}_{\text{luk}}}(u_0)\}.$$

Hence it suffices to show that $0.5 \leq m(y)$ for all $m \in \text{mod}_{\mathcal{M}_{\text{luk}}}(u_0)$, with equality holding for at least one $m$. So let $m \in \text{mod}_{\mathcal{M}_{\text{luk}}}(u_0)$. Then we have $0.75 \leq m(x)$, $0.75 \leq m(x \rightarrow y)$, and $0.25 \leq m(z)$. Unpacking the second constraint gives us $0.75 \leq 1 \barwedge (1 - m(x) + m(y))$ which leads to   $m(x) - 0.25 \leq m(y)$.

Since $0.75 \leq m(x)$ this gives us the desired $0.5 \leq m(y)$. Furthermore, we can obtain equality here by choosing $m_0 \in \text{mod}_{\mathcal{M}_{\text{luk}}}(u_0)$ such that $m_0(x) = 0.75, m_0(y) = 0.5$ and $m_0(z) = 0.25$. Hence $D_{\text{luk}}(u_0)(y) = 0.5$ as required. By similar reasoning we can also show $D_{\text{luk}}(u_0)(y \wedge z) = \min\{0.5, 0.25\} = 0.25$, i.e., we infer that the truth-degree of $y \wedge z$ is at least 0.25. So, by (*) above, we know $u_0 \sqcup (\neg(y \wedge z)/b)$ will be inconsistent for any $b > 0.75$.

### 2.1.3 Necessity Logic

Our final two examples show how the framework is also able to capture some types of non-truth-functional belief. The first of these, which corresponds to possibilistic logic [3], was described within this framework in [6]. For the semantics we take the set $\mathcal{M}_N$ of all *necessity functions* over $L_{\text{Prop}}$, i.e., the set of functions $n : L_{\text{Prop}} \to [0, 1]$ which satisfy, for all $\theta, \phi \in L$,

**(N1)** If $\theta \in Cn(\emptyset)$ then $n(\theta) = 1$ and $n(\neg\theta) = 0$.
**(N2)** If $(\theta \leftrightarrow \phi) \in Cn(\emptyset)$ then $n(\theta) = n(\phi)$.
**(N3)** $n(\theta \wedge \phi) = n(\theta) \wedge n(\phi)$.

We then take $D_N = J_{\mathcal{M}_N}$.

**Proposition 3 ([6])** $D_N$ *is logically compact.*

In this logic the notion of consistency is reducible to classical propositional consistency, in that a fuzzy base $u$ is $D_N$-consistent iff $Supp(u)$ is $Cn$-consistent. Also, if $u$ is consistent (and $\phi \notin Cn(\emptyset)$) then $D_N(u)(\phi)$ may be determined from the values given to those formulas which classically imply $\phi$ as follows:

$D_N(u)(\phi) = $
$\sup\{u(\theta_1) \wedge \ldots \wedge u(\theta_k) \mid \phi \in Cn(\{\theta_1, \ldots, \theta_k\})\}$

(If $\phi \in Cn(\emptyset)$ then clearly $D_N(u)(\phi) = 1$.) For example using the same fuzzy base $u_0$ as in the previous example we get $D_N(u_0)(y) = 0.75$ and $D_N(u_0)(y \wedge z) = 0.25$.

### 2.1.4 Probability Logic (Lower Envelopes)

Our last example is probabilistic. It is the logic of "lower envelopes" studied in [5].[2] This time we take as a semantics the set $\mathcal{M}_P$ of all *probability functions* over $L_{\text{Prop}}$, i.e., all functions $p : L_{\text{Prop}} \to [0, 1]$ which satisfy, for all $\theta, \phi \in L_{\text{Prop}}$,

**(P1)** If $\theta \in Cn(\emptyset)$ then $p(\theta) = 1$.
**(P2)** If $\neg(\theta \wedge \phi) \in Cn(\emptyset)$ then $p(\theta \vee \phi) = p(\theta) + p(\phi)$.

Then every "world" contains complete information of a random phenomena. We then take $D_P = J_{\mathcal{M}_P}$.

**Proposition 4 ([5, 6])** $D_P$ *is logically compact.*

A fuzzy base $u$ then gives a lower constraint for an unknown probability distribution. The deduction operator $D_P(u)$ improves the initial constraint. It is easy to see (using **(P2)**) that $D_P$ satisfies the property (*) mentioned in Sect. 2.1.2. A syntactic characterisation of $D_P$ may be found in [6]. For an example of an inference in this logic consider again the fuzzy base $u_0$ from the previous examples. Then $D_P(u_0)(y) = 0.5$, i.e., we infer that the probability of $y$ is at least 0.5. To see this, we have

$$D_P(u_0)(y) = \inf\{p(y) \mid p \in \text{mod}_{\mathcal{M}_P}(u_0)\}.$$

Hence it suffices to show $0.5 \leq p(y)$ for all $p \in \text{mod}_{\mathcal{M}_P}(u_0)$, with equality holding for at least one $p$. To see this, first note that, using the properties of probability functions, we get $p(x) = p(x \wedge y) + p(x \wedge \neg y)$ and $p(x \to y) = 1 - p(\neg(x \to y)) = 1 - p(x \wedge \neg y)$. Hence we may rewrite the first two constraints on $p$ as

$$0.75 \leq p(x \wedge y) + p(x \wedge \neg y) \text{ and } p(x \wedge \neg y) \leq 0.25.$$

The first constraint gives $0.75 - p(x \wedge \neg y) \leq p(x \wedge y)$. Then using this with the second constraint gives $0.5 \leq p(x \wedge y)$. Since $p(x \wedge y) \leq p(y)$ for any probability function we then get $0.5 \leq p(y)$ as required. We obtain equality by choosing any $p_0 \in \text{mod}_{\mathcal{M}_P}(u_0)$ such that $p_0(x \wedge \neg y) = p_0(\neg x \wedge \neg y) = 0.25$ and $p_0(x \wedge y) = 0.5$.

Note here that the answer for $D_P(u_0)(y)$ coincides with that for $D_{\text{luk}}(u_0)(y)$ in the Lukasiewicz example above. In general, though, the two deduction operators will give different results.[3] For example it can be shown that, in contrast to $D_{\text{luk}}(u_0)$, we get $D_P(u_0)(y \wedge z) = 0$.

## 3 FUZZY BASE REVISION

Now we have set up the basic framework we can state formally the question of revision we are interested in:

**Question.** Assume a fixed abstract fuzzy deduction system $(L, W, D)$ as background. Then given a fuzzy belief base $u$ (representing our current (fuzzy) information) and a pair $(\phi/a) \in L \times W$ (representing the new information that the truth-degree of $\phi$ is at least $a$), how should we determine $u \star (\phi/a)$ which represents the *revision* of $u$ to consistently incorporate the new information $(\phi/a)$?

The special case of crisp deduction systems is the case which is considered in the AGM framework. The idea there is to decompose the operation into two main steps. *First*, the initial (crisp) base $u$ is altered if necessary so as to "make room" for, i.e., become consistent with, the incoming crisp formula $\phi$. This is achieved by making $u$ deductively weaker (contraction). Here we should adhere to the principle of *minimal change*, according to which this weakening should be made as "small" as possible. *Then* the new formula is simply joined on to the result (expansion). In *partial meet revision* [1, 11] the idea is to focus for the first step on those subsets of $u$ which are consistent with $\phi$ and which are *maximal* with this property. Then, a certain number of the elements of this set are somehow selected as the "best" or "most preferred" and then their intersection is taken. The result of this intersection is then expanded by $\phi$. We would like to generalise this procedure to apply to an arbitrary abstract fuzzy deduction system. In other words we want to use the following procedure to obtain $u \star (\phi/a)$:

1. Form the family of *maximal fuzzy subsets of $u$ which are consistent with $(\phi/a)$*. We denote this family by $u \perp (\phi/a)$.[4]
2. Select a subset of these: $\gamma(u \perp (\phi/a)) \subseteq u \perp (\phi/a)$.
3. Form the meet of the elements of this subset: $\bigsqcap \gamma(u \perp (\phi/a))$.
4. Join $(\phi/a)$ to the result: $u \star (\phi/a) = (\bigsqcap \gamma(u \perp (\phi/a))) \sqcup (\phi/a)$.

---
[2] See also [6] for some more examples of "probability-like" logics within this framework.

[3] See also [9].

We now fill in the details of the above sketched procedure. First we formally define $u \perp (\phi/a)$:

**Definition 4** *Let $u \in \mathcal{F}(L)$ and $(\phi/a) \in L \times W$. Then $u \perp (\phi/a)$ is the set of elements of $\mathcal{F}(L)$ such that $u' \in u \perp (\phi/a)$ iff (i) $u' \sqsubseteq u$, (ii) $u' \sqcup (\phi/a)$ is consistent, and (iii) for all $u'' \sqsubseteq u$, if $u' \sqsubset u''$ then $u'' \sqcup (\phi/a)$ is inconsistent.*

Note in particular that if $u \sqcup (\phi/a)$ is consistent then $u \perp (\phi/a) = \{u\}$, while if $(\phi/a)$ is inconsistent then $u \perp (\phi/a) = \emptyset$. We need to know that if $(\phi/a)$ is consistent then $u \perp (\phi/a)$ is non-empty. In fact this is the main place where the property of logical compactness of $D$ is required. Under the additional assumption of Zorn's Lemma, it enables us to show the following:

**Proposition 5** *Let $v \in \mathcal{F}(L)$. If $v \sqsubseteq u$ and $v \sqcup (\phi/a)$ is consistent then there exists $w \in u \perp (\phi/a)$ such that $v \sqsubseteq w$.*

*Proof (Sketch).* First consider the set $X = \{u' \in \mathcal{F}(L) \mid v \sqsubseteq u' \sqsubseteq u, \ u' \sqcup (\phi/a) \text{ is consistent}\}$, partially ordered by $\sqsubseteq$. With the help of logical compactness, it can be shown that, for every (non-empty) totally-ordered subset $Y$ of $X$, the element $\bigsqcup Y$ is an upper-bound for $Y$ in $X$. (If $Y$ is empty then $v$ is an upper-bound for $Y$ in $X$.) Applying Zorn's Lemma, we then deduce the existence of a maximal element $w$ of $X$. It can then be shown that for any such $w$ we have both $w \in u \perp (\phi/a)$ and $v \sqsubseteq w$. □

Taking $v = u_\top$ in the above proposition gives us the desired non-emptiness for $u \perp (\phi/a)$. We now define selection functions.

**Definition 5** *Let $u \in \mathcal{F}(L)$. A selection function for $u$ is a function $\gamma$ such that for all $(\phi/a) \in L \times W$, (i) if $u \perp (\phi/a) \neq \emptyset$ then $\emptyset \neq \gamma(u \perp (\phi/a)) \subseteq u \perp (\phi/a)$, and (ii) if $u \perp (\phi/a) = \emptyset$ then $\gamma(u \perp (\phi/a)) = \{u\}$.*

Intuitively, selection functions reflect the *resistance to change* of the items of information in $u$. Given $u \in \mathcal{F}(L)$ and a selection function $\gamma$ for $u$ we then define a *revision operator* $\star_\gamma$ for $u$ as follows:

$$u \star_\gamma (\phi/a) = \left(\bigsqcap \gamma(u \perp (\phi/a))\right) \sqcup (\phi/a)$$

**Definition 6** *Let $u \in \mathcal{F}(L)$ and let $\star$ be an operator for $u$. Then $\star$ is an operator of* partial meet fuzzy base revision *(for $u$) iff $\star = \star_\gamma$ for some selection function $\gamma$ for $u$.*

The following proposition is reminiscent of the Harper Identity from crisp revision [4]. It is used later in the proof of Theorem 1.

**Proposition 6** *Let $\gamma$ be a selection function for $u$. Then $u \sqcap (u \star_\gamma (\phi/a)) = \bigsqcap \gamma(u \perp (\phi/a))$.*

*Proof (Sketch).* For the "$\sqsubseteq$" direction first note that $u \sqcap (u \star_\gamma (\phi/a)) = (u \sqcap (\bigsqcap \gamma(u \perp (\phi/a)))) \sqcup (u \sqcap (\phi/a))$ (using the distributivity of $\mathcal{F}(L)$). Hence it suffices to show both $u \sqcap (\bigsqcap \gamma(u \perp (\phi/a))) \sqsubseteq \bigsqcap \gamma(u \perp (\phi/a))$ and $u \sqcap (\phi/a) \sqsubseteq \bigsqcap \gamma(u \perp (\phi/a))$. The former clearly holds. The latter too if $(\phi/a)$ is inconsistent (since then $\gamma(u \perp (\phi/a)) = \{u\}$), while if $(\phi/a)$ is consistent it can be shown that $u \sqcap (\phi/a) \sqsubseteq u'$ for all $u' \in u \perp (\phi/a)$, which then suffices (since $\gamma(u \perp (\phi/a)) \subseteq u \perp (\phi/a)$). For the "$\sqsupseteq$" direction we need both $\bigsqcap \gamma(u \perp (\phi/a)) \sqsubseteq u$ and $\bigsqcap \gamma(u \perp (\phi/a)) \sqsubseteq (\bigsqcap \gamma(u \perp (\phi/a))) \sqcup (\phi/a)$. The latter clearly holds, while for the former we have $\gamma(u \perp (\phi/a)) \neq \emptyset$ and so, given $u' \in \gamma(u \perp (\phi/a))$, we have $\bigsqcap \gamma(u \perp (\phi/a)) \sqsubseteq u' \sqsubseteq u$ as required (since $u' \sqsubseteq u$ by definition of $u \perp (\phi/a)$). □

Thus $u \sqcap (u \star_\gamma (\phi/a))$ may be equated with the result of "contracting" $u$ to make room for the new item $(\phi/a)$.

### 3.1 EXAMPLES

Let us give an example of partial meet fuzzy base revision "in action" for each of the instantiations of the framework we gave in Sect. 2.1.

#### 3.1.1 Crisp Deduction Systems

For crisp deduction systems the operation reduces to the usual partial meet base revision from [11]. For example suppose $u = \{x, x \to y, z\}$ and suppose we receive the new information $\neg(y \wedge z)$ (it is understood that all the stated formulas have degree 1). Then we get

$$u \perp (\neg(y \wedge z)) = \{\{x, x \to y\}, \{x, z\}, \{x \to y, z\}\}.$$

---

[4]In the (crisp) belief revision literature the talk is usually (and equivalently) of the set of "maximal subsets which fail to imply $\neg\phi$", which is denoted by $u \perp \neg\phi$. We prefer the slightly different notation which does not refer to any connectives.

Suppose our selection function $\gamma$ selects the first two subsets above: $\gamma(u \perp (\neg(y \wedge z))) = \{\{x, x \to y\}, \{x, z\}\}$. Then we get $u \star_\gamma \neg(y \wedge z) = (\bigcap \gamma(u \perp (\neg(y \wedge z)))) \cup \{\neg(y \wedge z)\} = \{x, \neg(y \wedge z)\}$.

### 3.1.2 Łukasiewicz Fuzzy Logic

Suppose $u_0$ is given as in Sect. 2.1.2, i.e., $u_0 = \{(x/0.75), (x \to y/0.75), (z/0.25)\}$. Then suppose we receive the new information $(\neg(y \wedge z)/1)$, i.e., it is definitely *not* the case that $y$ and $z$ are true together. We know from the remark at the end of Sect. 2.1.2 that $u_0$ is inconsistent with this new information. In order to make $u_0$ consistent with $(\neg(y \wedge z)/1)$ we need to modify it so that $D_{\text{luk}}(u_0)(y \wedge z) = 0$. This can be achieved either by holding the truth-degrees of $x$ and $x \to y$ fixed while lowering that of $z$ to 0, or by holding the truth-degree of $z$ fixed and lowering that of either $x$ or $x \to y$ (or both) just enough to ensure $D_{\text{luk}}(u_0)(y) = 0$. Precisely, we can show that

$$u_0 \perp (\neg(y \wedge z)/1) = \{\{(x/0.75), (x \to y/0.75)\}\} \cup \{u' \sqsubseteq u_0 \mid 0.25 \leq u'(x), \\ u'(x \to y) = 1 - u'(x), \\ u'(z) = u_0(z)\}.$$

Suppose we prefer to keep the information item $(x/0.75)$, and that this is reflected by applying the selection function

$$\gamma(u_0 \perp (\neg(y \wedge z)/1)) = \{u' \in u_0 \perp (\neg(y \wedge z)/1) \mid \\ u'(x) = u_0(x)\}.$$

Then, using $u^*$ as shorthand for $u_0 \star_\gamma (\neg(y \wedge z)/1)$, we have $u^*(\neg(y \wedge z)) = 1$, while for $\theta \neq \neg(y \wedge z)$ we have

$$u^*(\theta) = \left[\bigsqcap \gamma(u_0 \perp (\neg(y \wedge z)/1))\right](\theta) \\ = \inf\{u'(\theta) \mid u' \in \gamma(u_0 \perp (\neg(y \wedge z)/1))\}.$$

Hence, as our final result we get $u_0 \star_\gamma (\neg(y \wedge z)/1) = \{(x/0.75), (x \to y/0.25), (\neg(y \wedge z)/1)\}$.

### 3.1.3 Necessity Logic

Let $u_0$ be as in the previous example and suppose we get the new information $(\neg(y \wedge z)/0.25)$. Then, since $Supp(u_0 \sqcup (\neg(y \wedge z)/0.25)) = \{x, x \to y, z, \neg(y \wedge z)\}$ is $Cn$-inconsistent we know $u_0 \sqcup (\neg(y \wedge z)/0.25)$ is inconsistent. Finding the fuzzy subsets of $u_0$ which are maximally consistent with $(\neg(y \wedge z)/0.25)$ essentially reduces to finding the crisp subsets of $Supp(u_0)$ which are maximally $Cn$-consistent with $\neg(y \wedge z)$:

$$u_0 \perp (\neg(y \wedge z)/0.25) = \{\{(x \to y/0.75), (z/0.25)\}, \\ \{(x/0.75), (z/0.25)\}, \\ \{(x/0.75), (x \to y/0.75)\}\}$$

Hence so far this doesn't look much different from the case of crisp deduction systems. The only difference is that now not all the formulas have degree 1. We have the option of using this extra expressiveness to actually help *define* a selection function, perhaps according to a principle that formulas with greater degrees should be kept whenever possible. Indeed this is the approach usually taken in works on belief revision within possibility theory such as [2]. For instance in the above example we could prefer to throw out the information item with the lowest degree, i.e., $(z/0.25)$. This would be reflected by using a selection function for $u_0$ such that:

$$\gamma(u_0 \perp (\neg(y \wedge z)/0.25)) = \{\{(x/0.75), (x \to y/0.75)\}\}$$

Then

$$\bigsqcap(\gamma(u_0 \perp (\neg(y \wedge z)/0.25))) = \gamma(u_0 \perp (\neg(y \wedge z)/0.25))$$

and so $u_0 \star_\gamma (\neg(y \wedge z)/0.25) = \{(x/0.75), (x \to y/0.75), (\neg(y \wedge z)/0.25)\}$.[5] We remark, however, that there is nothing to stop us from defining $\gamma$ independently of the degrees.[6]

### 3.1.4 Probability Logic (Lower Envelopes)

For a probabilistic example let us again use the base $u_0$ from earlier and suppose this time we get new information $(\neg y/0.75)$ which, since as we saw in Sect. 2.1.4 $D_P(u_0)(y) > 0.25$, is inconsistent $u_0$. Then it can be shown that

$$u_0 \perp (\neg y/0.75) = \{u' \sqsubseteq u_0 \mid 0.5 \leq u'(x), \\ u'(x \to y) = 1.25 - u'(x), \\ u'(z) = u_0(z)\}$$

Suppose our selection function $\gamma$ is defined by

$$\gamma(u \perp (\neg y/0.75)) = \{u' \in u \perp (\neg y/0.75) \mid 0.6 \leq u'(x)\}$$

reflecting a certain "level of security" behind the item of information $(x/0.75)$: we are not willing to choose any subset of $u_0$ in which the probability of $x$ falls below 0.6. Then, using $u^*$ now as shorthand for $u_0 \star_\gamma (\neg y/0.75)$ we have $u^*(\neg y) = 0.75$, while for

---

[5] For a related approach see [16].
[6] In fact the question of the precise nature of the relationship between degrees of confidence (i.e., truth-degrees for us) and degree of resistance to change is one of the open philosophical problems in belief revision recently posed by Hansson [12].

$\theta \neq \neg y$ we have

$$u^*(\theta) = \left[\bigsqcap \gamma(u_0 \perp (\neg y/0.75))\right](\theta)$$
$$= \inf\{u'(\theta) \mid u' \in \gamma(u_0 \perp (\neg y/0.75))\}.$$

Hence $u_0 \star_\gamma (\neg y/0.75) =$
$\{(x/0.6), (x \to y/0.5), (z/0.25), (\neg y/0.75)\}$.

## 4 CHARACTERISING PARTIAL MEET FUZZY BASE REVISION

In this section we axiomatically characterise the class of partial meet fuzzy base revision operators. It turns out that the class is characterised by the following five postulates, each of which generalises a postulate from the corresponding axiomatisation from the crisp case [11]. On the right we list the usual names.

(**F1**) $a \leq_W [u \star (\phi/a)](\phi)$ (Success)

(**F2**) $u \star (\phi/a)$ is consistent if $(\phi/a)$ is consistent (Consistency)

(**F3**) $u \star (\phi/a) \sqsubseteq u \sqcup (\phi/a)$ (Inclusion)

(**F4**) For all $\theta \in L, b \in W$, if $b \not\leq_W [u \star (\phi/a)](\theta)$ and $b \leq_W u(\theta)$ then there exists $u'$ such that $u \star (\phi/a) \sqsubseteq u' \sqsubseteq u \sqcup (\phi/a)$, $u'$ is consistent and $u' \sqcup (\theta/b)$ is inconsistent (Relevance)

(**F5**) If, for all $v \sqsubseteq u$, we have $v \sqcup (\phi/a)$ is consistent iff $v \sqcup (\phi'/a')$ is consistent, then $u \sqcap (u \star (\phi/a)) = u \sqcap (u \star (\phi'/a'))$ (Uniformity)

(**F1**) says that the revision is *successful*, i.e., that after revision by $(\phi/a)$, the formula $\phi$ is assigned a truth-degree of at least $a$. (**F2**) requires the result of revision to be consistent, provided the input is itself consistent. (**F3**) says that the revised base should not contain more information than that obtained by simply joining the original base with the new information. (**F4**) seeks to minimise unnecessary loss of information. Roughly, it expresses that if, for every consistent fuzzy base $u'$ lying between $u \star (\phi/a)$ and $u \sqcup (\phi/a)$, it is possible to raise the truth-degree of $\theta$ from $u'(\theta)$ to $b$ without incurring inconsistency, then there is no reason for the revised truth-degree of $\theta$ to fall below $b$. Finally for (**F5**), first note that $u \sqcap (u \star (\phi/a))$ can be understood as that information in $u$ which is retained in $u \star (\phi/a)$. Hence (**F5**) says that if two different inputs are consistent with precisely the same fuzzy subsets of $u$ then they remove the same information from $u$. We now give the main result of the paper, which generalises the characterisation given in [11] for crisp deduction systems.

**Theorem 1** *Let $u \in \mathcal{F}(L)$ and $\star$ be an operator for $u$. Then $\star$ is an operator of partial meet fuzzy base revision for $u$ iff $\star$ satisfies (**F1**)–(**F5**).*

*Remarks on the proof.* The proof is based on that of the special crisp case in [11]. The main difficulty arises from the unavailability in our more general case of the (Deduction) property. Also, it turns out that the only properties of $D$ which are needed are logical compactness and the following weakening of (Monotony):

If $v$ is consistent and $u \sqsubseteq v$ then $u$ is consistent.

This last remark also applies to propositions 5 and 6.[7]

The next proposition gives us some more rules which can be derived from (**F1**)–(**F5**).

**Proposition 7** *Let $u \in \mathcal{F}(L)$ and $\star$ be an operator for $u$ which satisfies (**F1**)–(**F5**). Then $\star$ also satisfies the following properties:*

(**F6**) If $u \sqcup (\phi/a)$ is consistent then
$$u \star (\phi/a) = u \sqcup (\phi/a)$$

(**F7**) If $u$ is consistent and $a \leq_W u(\phi)$ then
$$u \star (\phi/a) = u$$

(**F8**) If $u(\phi) \leq_W a$ then $[u \star (\phi/a)](\phi) = a$

(**F9**) If $(\phi/a)$ is inconsistent then
$$u \star (\phi/a) = u \sqcup (\phi/a)$$

(**F10**) If $D(\phi/a) = D(\phi'/a')$ then
$$u \sqcap (u \star (\phi/a)) = u \sqcap (u \star (\phi'/a'))$$

(**F6**) is the "vacuity" property which says that if the new information $(\phi/a)$ is consistent with the current information $u$, then the new base is formed by simply adding $(\phi/a)$ to $u$. As a consequence of this we get (**F7**), which says that if $u$ is consistent and $\phi$ is already explicitly assigned a truth-degree in $u$ of at least $a$ then revising by $(\phi/a)$ leaves the base unchanged. (**F8**) says that if $u(\phi) \leq_W a$ then $\phi$ is assigned a truth-degree in the new base of *precisely* $a$. For the common case when $W$ is linearly ordered, (**F7**) and (**F8**) together give:

If $u$ is consistent then $[u \star (\phi/a)](\phi) = u(\phi) \vee a$.

We remark, however, that it can be shown partial meet fuzzy base revision operators do not satisfy this property in general. (**F9**) states that if the new information is inconsistent then the new base is again formed by just adding it to the current information. Finally, (**F10**) says that revising by information which is "logically equivalent" removes the same information from $u$. The proof of Prop. 7 requires only the same properties of $D$ as Theorem 1, with one small exception: the derivation of (**F10**) requires all three of the (generalised) Tarski properties.

---

[7]For the crisp case, it is already noticed in [14, Sect. 3] that the only properties required of $D$ are (Compactness) and (Monotony) (which, in the presence of (Deduction), is actually equivalent to its above weakening).

## 5 CONCLUSION

We have considered the question of fuzzy belief base revision within Gerla's general framework for fuzzy logic. We have defined and axiomatised the operation of partial meet fuzzy base revision, which generalises the operation of partial meet base revision from the usual crisp case. The fact that we obtained this axiomatisation with such relatively weak restrictions shows on the one hand how the ideas of rational belief change are general enough to be applied to reasoning under vagueness or uncertainty. On the other hand, it confirms that the types of fuzzy systems covered by our abstract setting are indeed appropriate for modelling the human capacity of making conclusions from uncertain or vague premises. We have given some examples which show how the operation works in some specific instances of the framework, including those related to Łukasiewicz fuzzy logic and probability logic.

In this paper the question of base revision has been investigated from a very high position on the abstraction ladder, with only a handful of properties assumed of the basic primitives. We have shown that it is nevertheless possible to formulate basic properties of base revision operators. We would like to think of (**F1**)–(**F5**) as the absolute minimal *core* properties which any base revision operator should satisfy. However, as we move down the abstraction ladder, we fully expect to be able to say more. Furthermore, as the differences between the various instantiations of our abstract framework then come into focus, such as those between truth-functional logic and uncertainty calculi (e.g. probability logic), we also expect to be able to answer another important question: are there postulates suitable for revision in one setting which are unsuitable in another? This will be left for future work, as will the consideration of postulates which govern the revision of a base by different, but related inputs. What, for example (assuming we work in $L_{\text{Prop}}$), is the connection between $u \star (\phi/b)$ and $u \star (\theta \wedge \phi/b)$? Also in this category would be some property of *robustness*, i.e., the idea that small changes in the degree $a$ of the revision input $(\phi/a)$ should cause only small changes to $u\star(\phi/a)$ (particularly relevant if $W = [0,1]$). Probably the fulfillment of conditions like these by partial meet fuzzy base revision operators will require some restriction on the selection function $\gamma$. Some preliminary investigations into the latter suggest we get robustness if we additionally restrict to continuous truth-functional semantics. Finally we would also like to study *theory* revision in this framework.

**Acknowledgements**
R. B.'s work is supported by the DFG project "Computationale Dialektik". Thanks are due to J. B. Paris, J. Malinowski and the referees for some useful comments and suggestions.

## References

[1] C. Alchourrón, P. Gärdenfors and D. Makinson, On the logic of theory change: Partial meet contraction and revision functions, *Journal of Symbolic Logic* **50** (1985) 510–530.

[2] D. Dubois and H. Prade, A synthetic view of belief revision with uncertain inputs in the framework of possibility theory, *International Journal of Approximate Reasoning* **17**(2-3) (1997) 295–324.

[3] D. Dubois, J. Lang and H. Prade, Possibilistic logic, in: *Handbook of Logic in Artificial Intelligence and Logic Programming*, Vol. 3, Clarendon Press (1994).

[4] P. Gärdenfors, *Knowledge in Flux*, MIT Press (1988).

[5] G. Gerla, Inferences in probability logic, *Artificial Intelligence* **70** (1994) 33–52.

[6] G. Gerla, Probability-like functionals and fuzzy logic, *Journal of Mathematical Analysis and Application* **216** (1997) 438–465.

[7] G. Gerla, *Fuzzy Logic: Mathematical Tools for Approximate Reasoning*, Kluwer Academic Publishers (2001).

[8] P. Hájek, *Metamathematics of Fuzzy Logic*, Kluwer Academic Publishers (1998).

[9] P. Hájek, L. Godo and F. Esteva, Fuzzy logic and probability, in: *Proceedings of UAI'95*, (1995) 237–244.

[10] J. Halpern, Plausibility measures: A general approach for representing uncertainty, in: *Proceedings of IJCAI'01*, (2001) 1474–1483.

[11] S. O. Hansson, Reversing the Levi identity, *Journal of Philosophical Logic*, **22** (1992) 637–669.

[12] S. O. Hansson, Ten philosophical problems in belief revision, *Journal of Logic and Computation*, **13** (2003) 37–49.

[13] S. O. Hansson, *A Textbook of Belief Dynamics*, Kluwer Academic Publishers (1999).

[14] S. O. Hansson and R. Wassermann, Local change, *Studia Logica* **70**(1) (2002) 49–76.

[15] V. Novák, I. Perfilieva and J. Močkoř, *Mathematical Principles of Fuzzy Logic*, Kluwer Academic Publishers (1999).

[16] R. Witte, Fuzzy belief revision, in: *Proceedings of NMR'02* (2002) 311–320.